\documentclass{article}
\usepackage{spconf,amsmath,graphicx}
\usepackage{multicol}
\usepackage{cite}

\title{ARBITRARY STYLE TRANSFER USING GRAPH INSTANCE NORMALIZATION}
\name{Dongki Jung \qquad Seunghan Yang \qquad Jaehoon Choi \qquad Changick Kim}
\address{Korea Advanced Institute of Science and Technology (KAIST), Daejeon, Korea}
\begin{document}

\maketitle

\begin{abstract}
Style transfer is the image synthesis task, which applies a style of one image to another while preserving the content.
In statistical methods, the adaptive instance normalization (AdaIN) whitens the source images and applies the style of target images through normalizing the mean and variance of features.
However, computing feature statistics for each instance would neglect the inherent relationship between features, so it is hard to learn global styles while fitting to the individual training dataset.
In this paper, we present a novel learnable normalization technique for style transfer using graph convolutional networks, termed Graph Instance Normalization (GrIN).
This algorithm makes the style transfer approach more robust by taking into account similar information shared between instances. 
Besides, this simple module is also applicable to other tasks like image-to-image translation or domain adaptation.
\end{abstract}
\vspace{-1mm}
\section{Introduction}
\label{sec:intro}
\subsection{Neural Style Transfer}
\label{ssec:subhead}
Nowadays, Convolutional Neural Networks have been successfully used for many style transfer tasks \cite{huang2017arbitrary, chelaramani2018cross, xing2018portrait, yao2019photo}.
Style transfer can be divided into two types depending on the method of stylizing.
First, it is a statistical approach that takes into consideration the relationship between channels.
Gatys \emph{et al.}\cite{gatys2015neural, gatys2016image}, Huang and Belongie\cite{huang2017arbitrary}, and Li \emph{et al.}\cite{li2017universal} propose various statistical methods to transfer any images into artworks using the style of famous artists.
Gatys \emph{et al.}\cite{gatys2016image,gatys2015neural} and Li \emph{et al.}\cite{li2017universal} exploit the Gram matrix, which represents the covariance of style features’ channels.
On the other hand, AdaIN\cite{huang2017arbitrary} calculates means and standard deviations of the features along the channel dimension.
The second method is to stack convolutional layers and transmit styles through deep networks for image-to-image translation.
Zhu \emph{et al.}\cite{zhu2017unpaired}, Liu \emph{et al.}\cite{liu2017unsupervised} and Huang \emph{et al.}\cite{huang2018multimodal} employ generative adversarial networks \cite{goodfellow2014generative} to translate source images to target images.
Choi \emph{et al.}\cite{choi2019self} make more natural stylized outputs by combining statistic methods with the network approach.
However, we focus on the statistical approach to suggest simple modules that can be applied to various networks.

In previous works, Gatys \emph{et al.}\cite{gatys2015neural, gatys2016image} present reconstruction losses by dividing features into style and content through deep learning.
\begin{figure}[t]
    \centering
    \includegraphics[width=\linewidth]{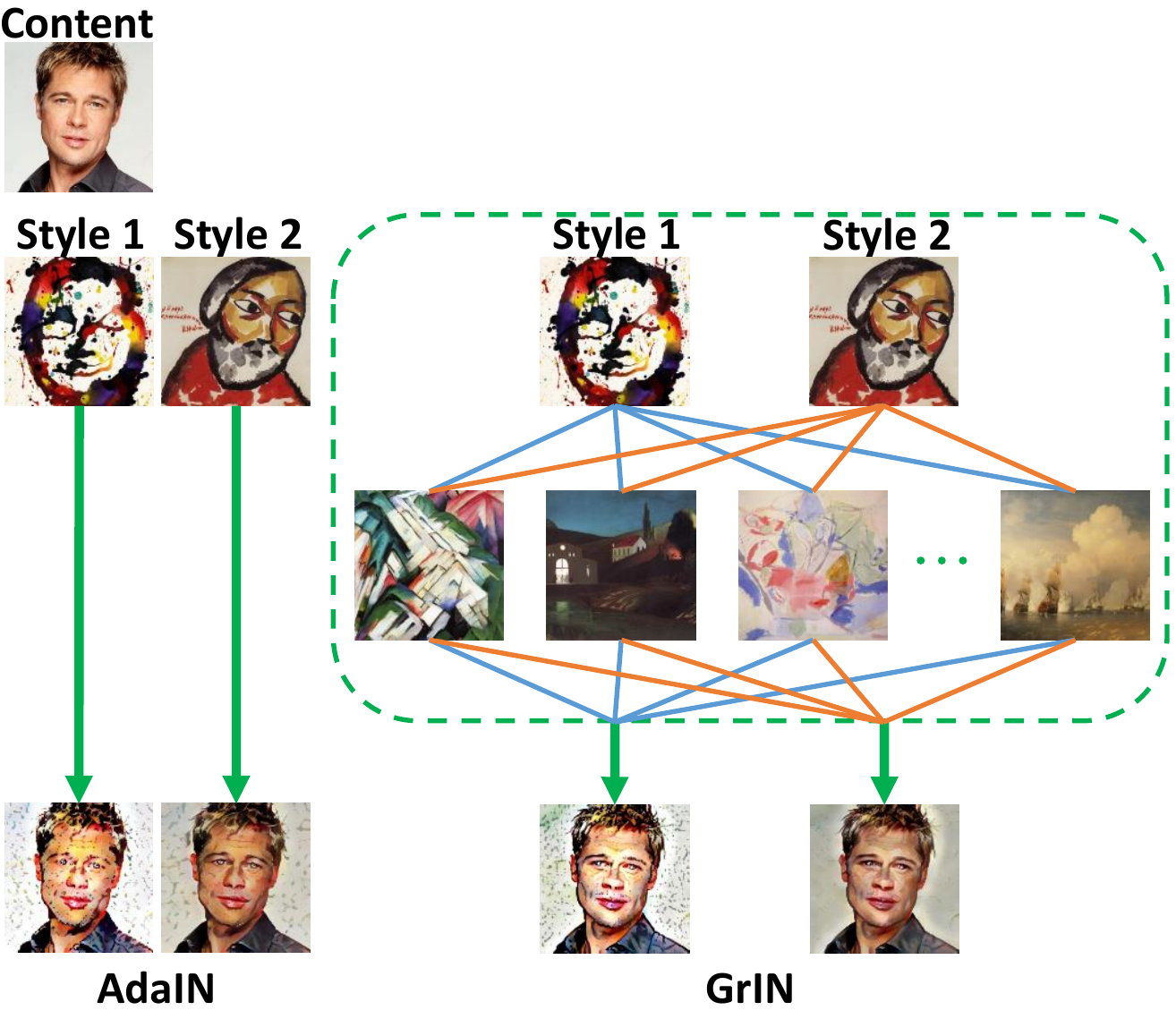}
    \vspace{-8mm}
    \caption{Schematic diagram of style transfer.
    Output images are at the bottom.
    The dotted line indicates a mini-batch of style images.
    AdaIN reflects only one target style to normalize statistics, but GrIN takes all the style images in the batch to learn general style information.}
    \vspace{-5mm}
    \label{fig:1}
\end{figure}
It has the disadvantage of slow optimization because content loss and style loss have to be updated iteratively whenever a new style has come.
AdaIN\cite{huang2017arbitrary} proposes methods that can transfer an arbitrary style, which operates with feed-forward networks in the real-time process.
Notably, it eliminates style information by directly normalizing feature statistics using perceptual loss \cite{johnson2016perceptual} and a statistical approach called adaptive instance normalization.
However, Nam and Kim\cite{nam2018batch} point out that instance normalization (IN) \cite{ulyanov2016instance} degrades the performance for other discriminative tasks.
This is because IN processes features independently and eliminates significant style variations of the content features.
Thus, they propose a way to apply style by mixing batch normalization (BN)\cite{ioffe2015batch} and IN using the gate parameter to remove the style information selectively.
Although it has the effect of BN partially to keep important style information, it cannot understand the one associated between features.
Therefore we design a new method with the graph convolutional layers to have the effect of graph smoothing on similar style features.
Figure \ref{fig:1} shows the difference between AdaIN and the proposed style transfer.
We propose a way to learn general styles with GCN by relating similarities between feature nodes to get style information, while AdaIN uses each feature information independently.
\vspace{-2mm}
\subsection{Graph Neural Networks}
\label{ssec:subhead}
\vspace{-1mm}

Recently, graph-based learning methods have been exploited in deep learning research.
Among them, the graph convolutional networks (GCN) \cite{kipf2016semi}, which are motivated by the first-order approximation of localized spectral filters on graphs \cite{hammond2011wavelets}, apply to various computer vision tasks, and they achieve state-of-the-art performance.
Specifically, GCN can adequately consider the label correlations \cite{chen2019multi}, data structure \cite{ma2019gcan}, and relatedness of the instances \cite{wu2019learning}.

In our work, we propose combining GCN with the normalization method to consider the correlation between style images.
In particular, to overcome the problem that IN ignores the feature relationship, we propose a normalization technique using graph convolutional networks, which can involve correlation in mini-batch samples with the adjacency matrix.
To the best of our knowledge, this is the first time to adapt the learnable graph layers into the style transfer.
Our method encourages the style transfer network to reduce some artifacts in output images and have insight into common style features by introducing simple graph convolution to the statistical approach.
Moreover, since a few graph layers are added for training and removed when inference, it maintains the high-speed advantage.

\begin{figure*}
    \centering
    \centerline{\includegraphics[width=\linewidth]{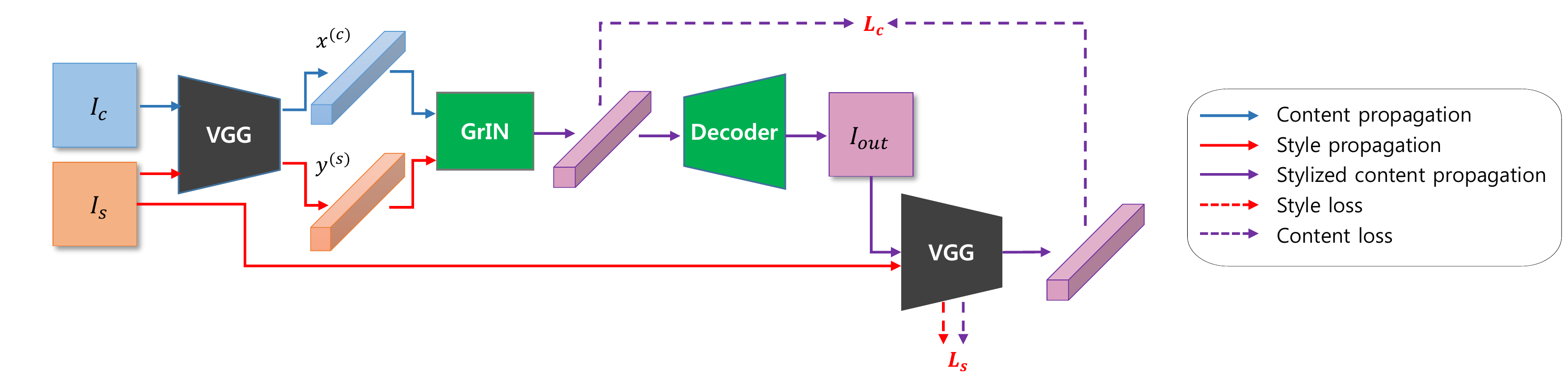}}
\vspace{-4mm}
\caption{Network architecture. $I_c$ and $I_s$ are content and style images, and $x^{(c)}$ and $y^{(s)}$ are the corresponding features by the fixed encoder, VGG-19. This network aims to learn the content and style information separately so that it can compose the stylized output, $I_{out}$. GrIN helps train the decoder more robustly by sharing the similar features' structural information. $L_c$ is the content loss, and $L_s$ is the style loss, mentioned in Section \ref{ssec:Training}.}
\vspace{-2mm}
\label{fig_2}
\end{figure*}

\begin{figure}
    \centering
    \centerline{\includegraphics[width=9.5cm]{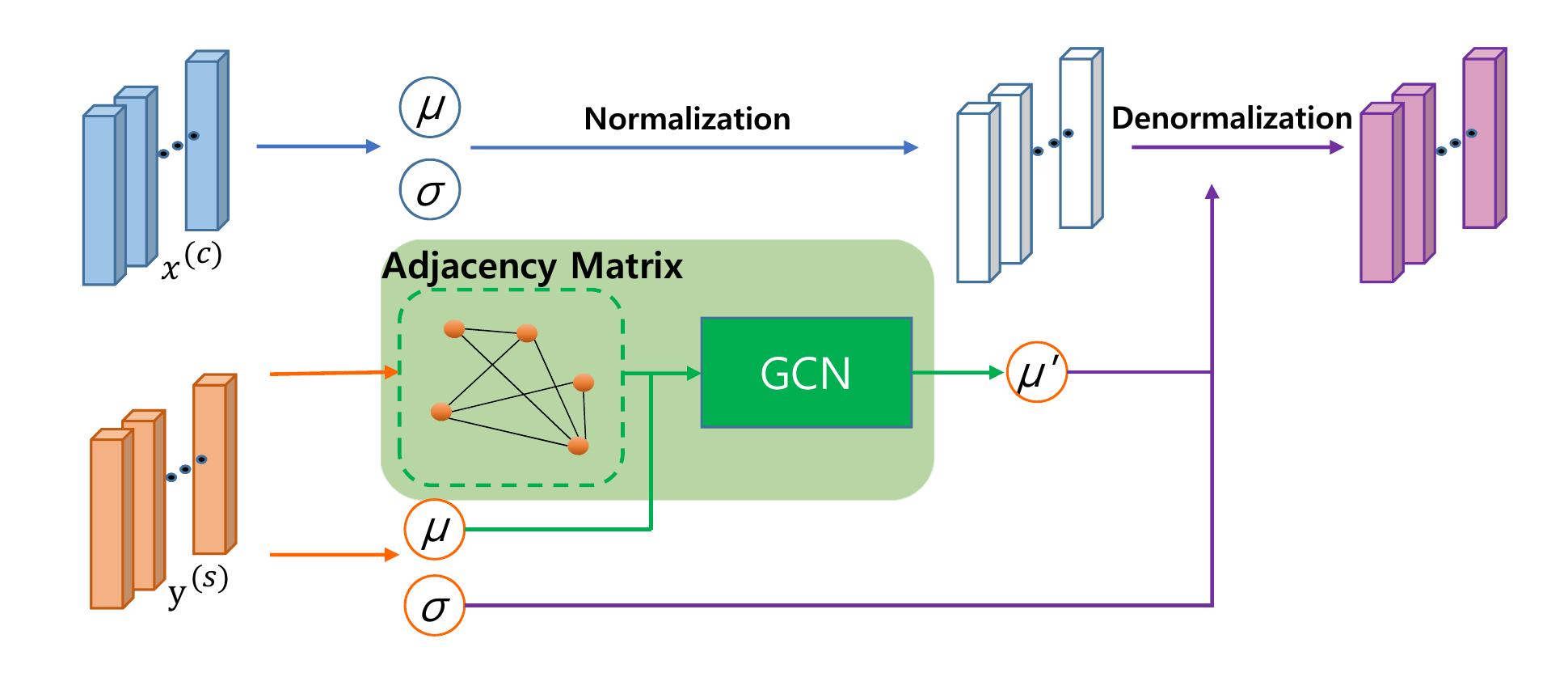}}
    \vspace{-2mm}
\caption{Green zone is our Graph Instance Normalization. This makes it possible to style by considering the
relationship between similar features. 
At inference time, our module is not used and $\mu'$ is equivalent to $\mu$.}
\vspace{-2mm}
\label{fig_3}
\end{figure}

\vspace{-1mm}
\section{Proposed method}
\label{sec:pagestyle}
\subsection{Normalization for Style Transfer}
\label{ssec:subhead}

\vspace{-1mm}
AdaIN is a statistical method that extracts the mean and standard deviation in each feature channel.
The statistics of content features $x\in \mathcal{R}^{N\times C\times H\times W}$ are calculated as follows:

\vspace{-1mm}
\begin{equation}
\mu_{nc}(x) = \frac{1}{HW}\sum_{h=1}^{H}\sum_{w=1}^{W}x_{nchw},
\end{equation}

\vspace{-4mm}
\begin{equation}
\sigma_{nc}^2(x) = \frac{1}{HW}\sum_{h=1}^{H}\sum_{w=1}^{W}(x_{nchw} - \mu_{nc}(x))^2.
\end{equation}
The same is true for $y$.
Note that $x_{nchw}$ is an element of the feature $x$, where $h$ and $w$ are the spatial location, $c$ is the channel index, and $n$ is the index of the sample in the mini-batch.
To translate the style, the mean $\mu(x)\in \mathcal{R}^{N\times C}$ and the standard deviation $\sigma(x)\in \mathcal{R}^{N\times C}$ normalize the content features $x$, and the normalized features are decomposed by the feature statistics of $y$:
\begin{equation}\label{adain}
    \text{AdaIN}(x,y) = \sigma(y)\frac{x-\mu(x)}{\sqrt{\sigma^2(x) + \epsilon}}+\mu(y),
\end{equation}
where $\epsilon$ is a small number added to avoid division by zero.
As you can see from the above equations, AdaIN processes normalization of each node independently, which removes all style information without considering the importance of style variations\cite{nam2018batch}.
Thus, we use the adjacency matrix and graph layers to normalize features on the batch dimension.
If style features in a mini-batch are similar, they are conducted together, preserving the common style information.
In Eq. \ref{adain}, the standard deviation $\sigma(y)$ is a scale, and the mean $\mu(y)$ is a bias.
In other words, the standard deviation vectors reflect the overall style, while the mean vectors deal with more style details.
Changing the standard deviation is undesirable because it has a risk of transforming the entire style.
For this reason, we correlate only mean vectors to normalize similar styles by the graph layers.
The mean vectors of the style features are normalized together properly in the batch through the structural information, eliminating unnecessary style variance that can be noise.
In contrast, standard deviation vectors are precluded from graph smoothing to preserve the global style with suppressing the distortion.
\vspace{-2mm}

\subsection{GCN for Style Transfer}
\label{ssec:subhead}

To consider relationships between style images, we exploit graph convolutional networks \cite{kipf2016semi}, which are motivated by localized spectral filters on graphs \cite{hammond2011wavelets}.
To find the structural information of features in the mini-batch, we set one style for one node.
Here, we describe the equation of localized spectral filters:
\begin{equation}
{g}_{\theta}\star x = U{g}_{\theta}{U}^{*}x,
\end{equation}
where $U$ is the matrix of eigenvectors from the normalized graph Laplacian with a learnable filter ${g}_{\theta} = diag(\theta)$ in the Fourier domain. 
$*$ denotes the transpose operation.
GCN limits spectral convolutions on graphs as the first order neighborhood layer-wise convolution operation.
Therefore, only the related nodes are allowed to be operated.
Each GCN layer is described as follows:

\vspace{-1mm}
\begin{equation}
    X^{(i+1)} = {\tilde{D}}^{-\frac{1}{2}}\tilde{A}{\tilde{D}}^{-\frac{1}{2}}X^{(i)}{\theta}^{(i)},
\end{equation}
where $X^{(i)}\in \mathcal{R}^{N\times F}$ and $X^{(i+1)}\in \mathcal{R}^{N\times F'}$ are the $F$-dimensional input and $F'$-dimensional output graph signal matrix with N nodes, respectively, and ${\theta}^{(i)} \in \mathcal{R}^{F\times F'}$ is a learnable weight kernel with $\tilde{A} = A + I_{N}$ and $\tilde{D}_{ii} = \sum_{j}\tilde{A}_{ij}$.

Using the above equations, the matrix of nodes is composed of the features from the encoder:
\begin{equation}
X = \text{Encoder}(X_{batch}),
\end{equation}
where $X_{batch}$ is a mini-batch of samples. 
Next, we resize the feature $X\in \mathcal{R}^{N\times C\times H\times W}$ into $X'\in \mathcal{R}^{N\times CHW}$, which is a 2-dimensional matrix:
\begin{equation}
\tilde{A} = X'X'^T,
\end{equation}
where $\tilde{A}\in \mathcal{R}^{N\times N}$ notices similar degrees between features according to their structural information.
The mean vectors obtained through instance normalization are processed to the next step with the graph layers:

\vspace{-1mm}
\begin{equation}
\mu_c = (\mu_{1c}, \mu_{2c},\;\cdot \cdot \cdot\;,\;\mu_{nc})^T,
\end{equation}

\vspace{-1mm}
\begin{equation}
\mu_c' = \tilde{D}^{-\frac{1}{2}}\tilde{A}\tilde{D}^{-\frac{1}{2}}\mu_c\theta,
\end{equation}
where $\theta$ means the learnable weight of the graph layer.
 
As a result, instance-normalized content features are decomposed by the style feature statistics processed through GCN,
\vspace{-2mm}
\begin{equation}
\text{GrIN}(x,y) = \sigma(y)\frac{x-\mu(x)}{\sqrt{\sigma^2(x) + \epsilon}}+\mu(y)'.
\end{equation}

Graph layers take into account the relationship between neighboring nodes.
The more layers are added to our network, the smoother it becomes because graph layers filter the related nodes together.
GrIN takes each style feature in the mini-batch as a node to produce the adjacency matrix, which is the degree of similarity between them.
Consequently, GrIN can normalize the style details using the adjacency matrix.
This helps learn the general characteristics by reducing the risk of overfitting on the training dataset.

\vspace{-4mm}
\section{Experiments}
\label{sec:typestyle}
\vspace{-2mm}
Figure \ref{fig_2} shows an overview of the whole network, and the detail of GrIN is shown in Fig. \ref{fig_3}.

\vspace{-3mm}
\subsection{Settings}
\label{ssec:subhead}
\vspace{-1mm}
We used content images from MS-COCO\cite{lin2014microsoft}, and employed style images from WikiArt\cite{nichol2016painter}.
Each dataset has about 80,000 training samples. With a pre-trained and fixed VGG-19 encoder\cite{simonyan2014very}, we trained our decoder with the Adam optimizer\cite{kingma2014adam}. We stacked two graph convolutional layers and had a batch size of 16 content-style image pairs to make a sufficient graph smoothing effect. We resized both images to 512$\times$512 resolution, then cropped random regions of size 256$\times$256 with preserving the aspect ratio.

\vspace{-2mm}
\subsection{Training}
\label{ssec:Training}
\vspace{-1mm}
To prove the performance of GCN, we follow similar learning schemes with AdaIN. We train our network using the loss function,

\vspace{-5mm}
\begin{equation}
L = L_c + \lambda L_s,
\end{equation}
which is a weighted sum of the content loss $L_c$ and the style loss $L_s$ with the style loss weight $\lambda$. We select lambda as 10 in the experiments. The content loss is the Euclidean distance between the target features and the style transferred output features. The target feature $t$ is the GrIN output:

\vspace{-1mm}
\begin{equation}
    t = \text{GrIN}(x, y),
\end{equation}
\begin{equation}
\newcommand\norm[1]{\left\lVert#1\right\rVert}
    L_c = \norm{\text{Encoder}(\text{Decoder}(t)) – t}_2^2.
\end{equation}

The style loss is composed of the mean and standard deviation of the original style's feature and the stylized output's:
\vspace{-2mm}
\begin{equation}
\newcommand\norm[1]{\left\lVert#1\right\rVert}
\begin{aligned}
L_s = \sum_{i=1}^{L}\norm{\mu(\phi_i(\text{Decoder}(t))) - \mu(\phi_i(y))}_2^2 \\+\sum_{i=1}^{L}\norm{\sigma(\phi_i(\text{Decoder}(t))) - \sigma(\phi_i(y))}_2^2,
\end{aligned}
\end{equation}
where each $\phi_i$ is a feature of the VGG-19 layer. We use 4 features which are \textit{relu1\_1, relu2\_2, relu3\_1}, and \textit{relu4\_1}.


\vspace{-1mm}
\section{Results and Analysis}
\label{sec:subhead}
\vspace{-2mm}

We exploit graph layers to improve the quality of output images. 
This can be done by graph smoothing the mean vectors, which represent the detail of style in the transfer.
For the test, graph layers are excluded for style transfer to be performed on a single image and emphasize the detail information of that image.
The network is stable even without the GCN since the standard deviation that is responsible for the overall style nuance does not pass graph layers during the training.

\begin{figure}[p]
    \centering
    \includegraphics[width=\linewidth]{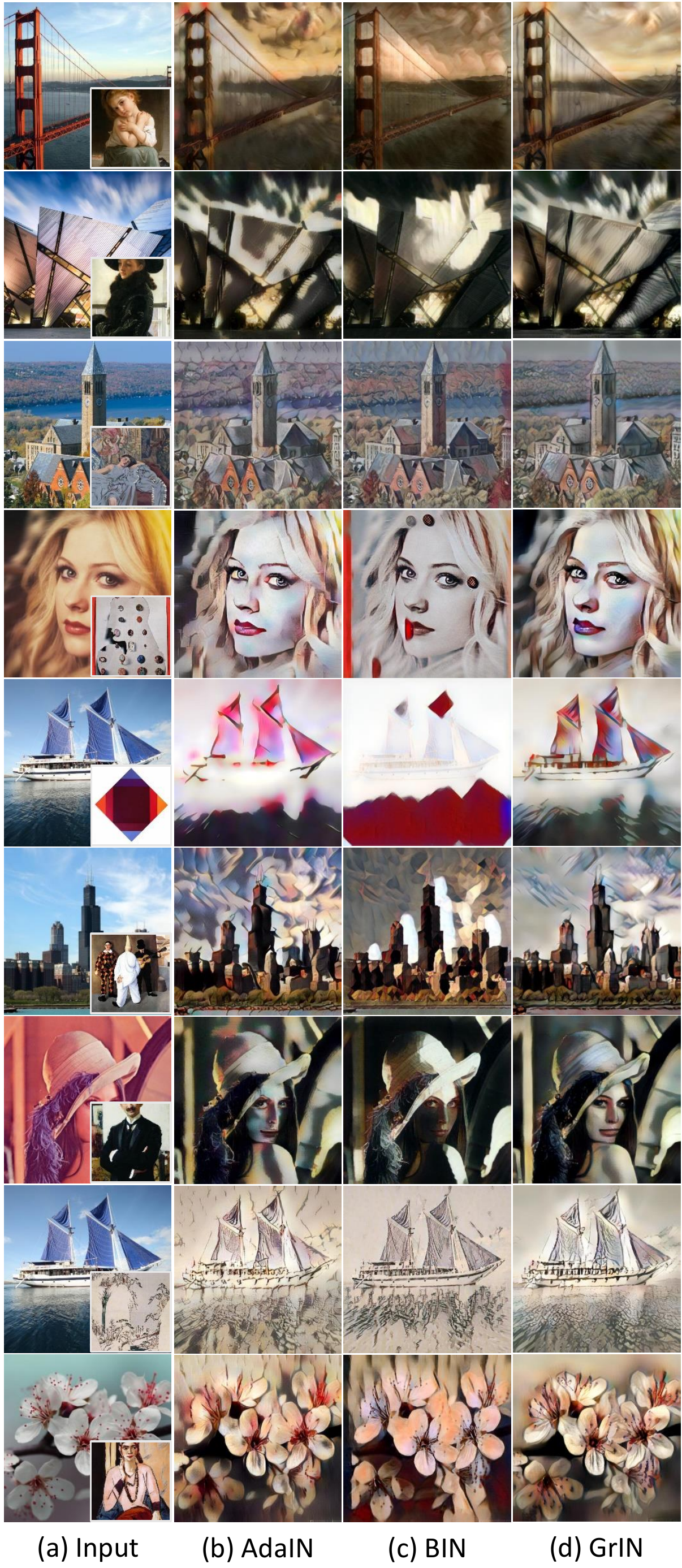}
    \caption{(a) shows the content image and the style image.
    (b), (c) and (d) are the results of Huang and Belongie\cite{huang2017arbitrary}, Nam and Kim\cite{nam2018batch}, and ours, respectively.
    The textual errors and the wash-out artifacts are reduced in ours.
    All the tested images are not shown to our network during training.
    }
    \label{fig:result_image}
\end{figure}

Figure \ref{fig:result_image} shows output images of our method and other style transfer algorithms.
In AdaIN, wash-out artifacts\cite{jamrivska2015lazyfluids} and textual errors are found frequently.
Although BIN shows good results for some styles while maintaining content information well, it simply stylizes colors on content images rather than understanding them.
As a result, like AdaIN, wash-out artifacts exist, and unintentional content information of style images appears sometimes.
However, GrIN considers similar style features together using an adjacency matrix.
Consequently, general style features can be learned by finding the common property among similar style images.
Since the general features help the transfer stylize appropriately to any arbitrary inputs during the test, GrIN can effectively transfer styles without noises and preserve the content images.


GrIN is a simple module with a few graph layers to learn general styles by the relationship between other style features.
Therefore, it can also be applied to image-to-image translation problems\cite{huang2018multimodal} or domain adaptation tasks\cite{choi2019self} based on AdaIN.
In future work, we will make use of our algorithm in those networks to improve the quality of stylized images.

\section{CONCLUSION}
\label{sec:majhead}

In this paper, we have designed a novel architecture, named GrIN, to learn the general styles of images. 
We integrate graph layers into AdaIN and modify the normalization scheme considering the mean as a bias term to overcome the inherent problem of instance normalization that cannot view the relationship between the features.
To the best of our knowledge, this is the first time to apply GCN to the task of style transfer.
The experimental result images show that GrIN produces more natural outputs than previous methods since graph convolutional networks induce to learn general styles by introducing the correlation between features.

\vspace{6mm}
\noindent \textbf{Acknowledgements} This work was partly supported by Institute of Information \& Communications Technology Planning \& Evaluation(IITP) grant funded by the Korea government(MSIT) (2017-0-01772. Development of QA system for video story understanding to pass Video Turing Test) and Institute of Information \& Communications Technology Planning \& Evaluation(IITP) grant funded by the Korea government(MSIT) (2017-0-01781.Data Collection and Automatic Tuning System Development for the Video Understanding)

\bibliographystyle{IEEEbib}
\bibliography{strings,refs}

\begin{thebibliography}{10}

\bibitem{huang2017arbitrary}
Xun Huang and Serge Belongie,
\newblock ``Arbitrary style transfer in real-time with adaptive instance
  normalization,''
\newblock in {\em ICCV}, 2017, pp. 1501--1510.

\bibitem{chelaramani2018cross}
Sahil Chelaramani, Abhishek Jha, and Anoop~M Namboodiri,
\newblock ``Cross-modal style transfer,''
\newblock in {\em ICIP}, 2018, pp. 2157--2161.

\bibitem{xing2018portrait}
Yeli Xing, Jiawei Li, Tao Dai, Qingtao Tang, Li~Niu, and Shu-Tao Xia,
\newblock ``Portrait-aware artistic style transfer,''
\newblock in {\em ICIP}, 2018, pp. 2117--2121.

\bibitem{yao2019photo}
Xu~Yao, Gilles Puy, and Patrick P{\'e}rez,
\newblock ``Photo style transfer with consistency losses,''
\newblock in {\em ICIP}, 2019, pp. 2314--2318.

\bibitem{gatys2015neural}
Leon~A Gatys, Alexander~S Ecker, and Matthias Bethge,
\newblock ``A neural algorithm of artistic style,''
\newblock {\em arXiv preprint arXiv:1508.06576}, 2015.

\bibitem{gatys2016image}
Leon~A Gatys, Alexander~S Ecker, and Matthias Bethge,
\newblock ``Image style transfer using convolutional neural networks,''
\newblock in {\em CVPR}, 2016, pp. 2414--2423.

\bibitem{li2017universal}
Yijun Li, Chen Fang, Jimei Yang, Zhaowen Wang, Xin Lu, and Ming-Hsuan Yang,
\newblock ``Universal style transfer via feature transforms,''
\newblock in {\em NeurIPS}, 2017, pp. 386--396.

\bibitem{zhu2017unpaired}
Jun-Yan Zhu, Taesung Park, Phillip Isola, and Alexei~A Efros,
\newblock ``Unpaired image-to-image translation using cycle-consistent
  adversarial networks,''
\newblock in {\em ICCV}, 2017, pp. 2223--2232.

\bibitem{liu2017unsupervised}
Ming-Yu Liu, Thomas Breuel, and Jan Kautz,
\newblock ``Unsupervised image-to-image translation networks,''
\newblock in {\em NeurIPS}, 2017, pp. 700--708.

\bibitem{huang2018multimodal}
Xun Huang, Ming-Yu Liu, Serge Belongie, and Jan Kautz,
\newblock ``Multimodal unsupervised image-to-image translation,''
\newblock in {\em ECCV}, 2018, pp. 172--189.

\bibitem{goodfellow2014generative}
Ian Goodfellow, Jean Pouget-Abadie, Mehdi Mirza, Bing Xu, David Warde-Farley,
  Sherjil Ozair, Aaron Courville, and Yoshua Bengio,
\newblock ``Generative adversarial nets,''
\newblock in {\em NeurIPS}, 2014, pp. 2672--2680.

\bibitem{choi2019self}
Jaehoon Choi, Taekyung Kim, and Changick Kim,
\newblock ``Self-ensembling with gan-based data augmentation for domain
  adaptation in semantic segmentation,''
\newblock in {\em ICCV}, 2019, pp. 6830--6840.

\bibitem{johnson2016perceptual}
Justin Johnson, Alexandre Alahi, and Li~Fei-Fei,
\newblock ``Perceptual losses for real-time style transfer and
  super-resolution,''
\newblock in {\em ECCV}, 2016, pp. 694--711.

\bibitem{nam2018batch}
Hyeonseob Nam and Hyo-Eun Kim,
\newblock ``Batch-instance normalization for adaptively style-invariant neural
  networks,''
\newblock in {\em NeurIPS}, 2018, pp. 2558--2567.

\bibitem{ulyanov2016instance}
Dmitry Ulyanov, Andrea Vedaldi, and Victor Lempitsky,
\newblock ``Instance normalization: The missing ingredient for fast
  stylization,''
\newblock {\em arXiv preprint arXiv:1607.08022}, 2016.

\bibitem{ioffe2015batch}
Sergey Ioffe and Christian Szegedy,
\newblock ``Batch normalization: Accelerating deep network training by reducing
  internal covariate shift,''
\newblock {\em arXiv preprint arXiv:1502.03167}, 2015.

\bibitem{kipf2016semi}
Thomas~N Kipf and Max Welling,
\newblock ``Semi-supervised classification with graph convolutional networks,''
\newblock {\em arXiv preprint arXiv:1609.02907}, 2016.

\bibitem{hammond2011wavelets}
David~K Hammond, Pierre Vandergheynst, and R{\'e}mi Gribonval,
\newblock ``Wavelets on graphs via spectral graph theory,''
\newblock {\em Applied and Computational Harmonic Analysis}, vol. 30, no. 2,
  pp. 129--150, 2011.

\bibitem{chen2019multi}
Zhao-Min Chen, Xiu-Shen Wei, Peng Wang, and Yanwen Guo,
\newblock ``Multi-label image recognition with graph convolutional networks,''
\newblock in {\em CVPR}, 2019, pp. 5177--5186.

\bibitem{ma2019gcan}
Xinhong Ma, Tianzhu Zhang, and Changsheng Xu,
\newblock ``Gcan: Graph convolutional adversarial network for unsupervised
  domain adaptation,''
\newblock in {\em CVPR}, 2019, pp. 8266--8276.

\bibitem{wu2019learning}
Jianchao Wu, Limin Wang, Li~Wang, Jie Guo, and Gangshan Wu,
\newblock ``Learning actor relation graphs for group activity recognition,''
\newblock in {\em CVPR}, 2019, pp. 9964--9974.

\bibitem{lin2014microsoft}
Tsung-Yi Lin, Michael Maire, Serge Belongie, James Hays, Pietro Perona, Deva
  Ramanan, Piotr Doll{\'a}r, and C~Lawrence Zitnick,
\newblock ``Microsoft coco: Common objects in context,''
\newblock in {\em ECCV}, 2014, pp. 740--755.

\bibitem{nichol2016painter}
K~Nichol,
\newblock ``Painter by numbers, wikiart,'' 2016.

\bibitem{simonyan2014very}
Karen Simonyan and Andrew Zisserman,
\newblock ``Very deep convolutional networks for large-scale image
  recognition,''
\newblock {\em arXiv preprint arXiv:1409.1556}, 2014.

\bibitem{kingma2014adam}
Diederik~P Kingma and Jimmy Ba,
\newblock ``Adam: A method for stochastic optimization,''
\newblock {\em arXiv preprint arXiv:1412.6980}, 2014.

\bibitem{jamrivska2015lazyfluids}
Ond{\v{r}}ej Jamri{\v{s}}ka, Jakub Fi{\v{s}}er, Paul Asente, Jingwan Lu, Eli
  Shechtman, and Daniel S{\`y}kora,
\newblock ``Lazyfluids: appearance transfer for fluid animations,''
\newblock {\em ACM Transactions on Graphics (TOG)}, vol. 34, no. 4, pp. 1--10,
  2015.

\end{thebibliography}

\end{document}